\documentclass[a4paper]{svproc}

\usepackage{graphicx}%
\usepackage{multirow}%
\usepackage{amsmath,amssymb,amsfonts}%
\usepackage{mathrsfs}%
\usepackage[title]{appendix}%
\usepackage{xcolor}%
\usepackage{textcomp}%
\usepackage{manyfoot}%
\usepackage{booktabs}%
\usepackage{algorithm}%
\usepackage{algorithmicx}%
\usepackage{algpseudocode}%
\usepackage{url}

\usepackage{subfig}
\usepackage{listings}%
\usepackage{caption}
\usepackage{hyperref}

\usepackage[font=small,labelfont=bf]{caption}
\usepackage{blindtext}

\graphicspath{{./images/}}

\begin{document}

\mainmatter
\title{Trailblazer: Learning off-road costmaps for long range planning}

\author{Kasi Viswanath \inst{1}, Felix Sanchez \inst{2}, Timothy Overbye \inst{2}, Jason M. Gregory \inst{2}, Srikanth Saripalli \inst{1}}
\authorrunning{Viswanath et al.}

\institute{Department of Mechanical Engineering, Texas A$\&$M University, USA\\
\email{\{kasiv, ssaripalli\}@tamu.edu}\\
\and
DEVCOM Army Research laboratory, Adelphi, MD, USA\\
\email{\{felix.a.sanchez12.ctr, timothy.j.overbye.ctr, jason.m.gregory1.civ\}@army.mil}}

\maketitle

\begin{abstract}
Autonomous navigation in off-road environments remains a significant challenge in field robotics, particularly for Unmanned Ground Vehicles (UGVs) tasked with search and rescue, exploration, and surveillance. Effective long-range planning relies on the integration of onboard perception systems with prior environmental knowledge, such as satellite imagery and LiDAR data. This work introduces Trailblazer, a novel framework that automates the conversion of multi-modal sensor data into costmaps, enabling efficient path planning without manual tuning. Unlike traditional approaches, Trailblazer leverages imitation learning and a differentiable A* planner to learn costmaps directly from expert demonstrations, enhancing adaptability across diverse terrains. The proposed methodology was validated through extensive real-world testing, achieving robust performance in dynamic and complex environments, demonstrating Trailblazer's potential for scalable, efficient autonomous navigation. \textcolor{blue}{\url{https://github.com/unmannedlab/Trailblazer}}     
\end{abstract}
\section{Introduction}
Unmanned Ground Vehicles (UGVs) are crucial across missions such as search and rescue, environmental exploration, and surveillance in challenging off-road terrains. While recent advancements have enabled autonomous systems to achieve long-range navigation in structured on-road environments, off-road mobility remains constrained by dynamically changing terrain properties, unexplored pathways, and sensor uncertainty. 

Though onboard sensors allow basic navigation, pre-loaded environmental data significantly boosts both safety and efficiency. This knowledge helps systems identify and avoid risky areas—such as unstable ground or steep slopes—while prioritizing safer routes. Without this understanding, robots rely solely on real-time sensing, increasing the likelihood of errors such as selecting unsafe paths. Integrating prior environmental information enables smarter, more reliable decision-making in unpredictable outdoor settings.

Overhead terrain data is a valuable source of prior environmental knowledge, especially when a vehicle encounters a site for the first time. These data sources include aerial or satellite imagery, digital elevation maps, and 3D LiDAR scans. Typically, this data is processed into 2D feature grids, which are then converted into traversal costs to guide autonomous navigation. While this approach has been effective in enhancing vehicle planning \cite{Vandapel-2003-8764}\cite{machines11080807}, it traditionally relies on hand-tuned cost functions, which are prone to bias. 

Previous work, \cite{6284849} proposed an approach to estimate a cost function for path planning using human demonstrations. This algorithm makes use of overhead data sources like satellite images and aerial LiDAR scans, processing them directly without going through the usual feature extraction steps. To manage the complexities of different scenes, linear regressors were applied. However, the absence of feature extraction can lead to a loss of important information. This may hinder the algorithm's effectiveness, especially in more complicated environments.

Recent advancements have introduced deep reinforcement learning \cite{hdif2023}\cite{9812238} and Vision Language Models \cite{huang23vlmaps} to optimize costmaps through human demonstrations, reducing manual intervention. However, many of these methods still depend on pre-trained convolutional neural networks (CNNs) trained on manually crafted costmaps, further refined using reinforcement learning algorithms.

Trailblazer offers a novel alternative by directly converting multi-modal sensor data into costmaps. Instead of relying on pre-trained models or manual annotations, Trailblazer learns cost-map generation from expert driven demonstrations. This approach simplifies the process, eliminates manual tuning, and enhances the system’s adaptability and efficiency in real-world scenarios for global planning.

\section{Methodology}
Trailblazer methodology consists of two stages: (i) data collection and pre-processing of satellite imagery and LiDAR data; (ii) Trailblazer framework and training for feature extraction and path planning.

\subsection{Data collection and pre-processing}
The Trailblazer architecture utilizes satellite/aerial imagery and airborne LiDAR point cloud data as primary inputs. The Trailblazer architecture makes use of satellite and aerial images, along with airborne LiDAR point cloud data, as its main sources of information. For this study, the data comes mainly from two reliable sources: the United States Geological Survey (USGS)\cite{USGSLidarExplorer} and the Environmental Systems Research Institute (ESRI), specifically through ArcGIS servers \cite{arcgis}. ESRI provides high-quality satellite images that can achieve a spatial resolution of up to 30 centimeters per pixel. On the other hand, the USGS LiDAR Explorer offers point cloud data with a density that goes beyond 8 points for every square meter.
\begin{figure}[ht]
    \centering    \includegraphics[width=\textwidth, height = 6 cm]{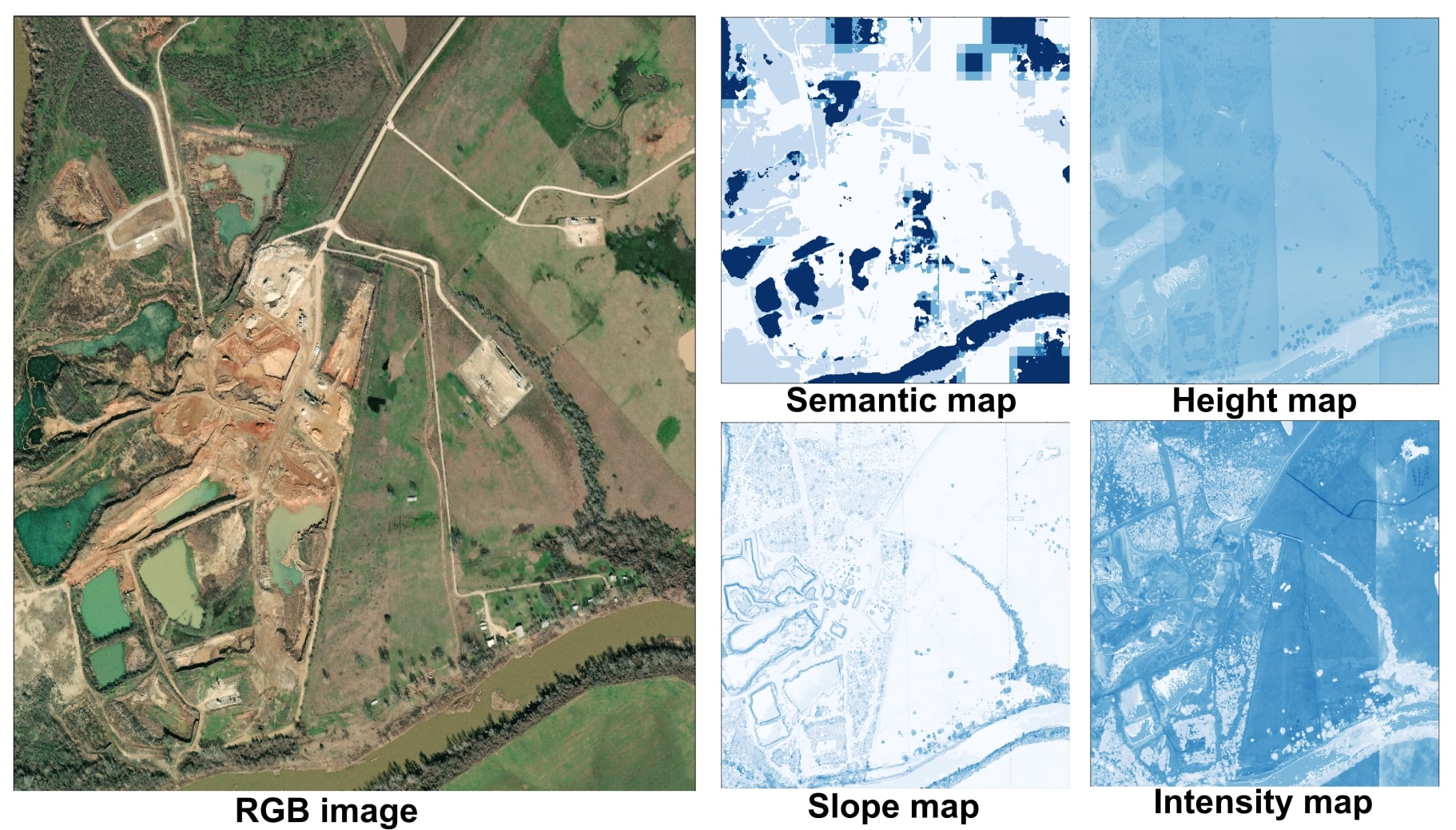}
    \caption{Inputs to Trailblazer from Texas A\&M RELLIS test site. The satellite image is of 30 $cm per pixel$ resolution used to generate semantic map from SegFormer. Overhead LiDAR data is used to generate the height, slope and intensity maps. The RELLIS test site covers an area of $1.5 km * 1.5 km$. }
    \label{fig:input}
\end{figure}
From the satellite image and aerial LiDAR data, we compute four maps as input for Trailblazer. The four maps consists of (i) Semantic segmentation, (ii) Height map, (iii) Slope profile (iv) Intensity map as shown in Figure \ref{fig:input}.

Semantic projection is derived using SegFormer\cite{xie2021segformer} on satellite imagery, classifying pixels by an ascending risk factor, from traversable to water. Segmented masks are projected onto LiDAR point clouds and mapped onto a grid. For each grid cell, the highest class index is assigned as its value. Geometric characteristics are extracted from LiDAR data. An average height map is created by calculating the average height of all points in each grid cell. LiDAR intensity is also analyzed on a per-cell basis. The average intensity per grid cell helps distinguish terrain types (e.g., trees, vegetation, barren land) based on surface characteristics. LiDAR intensity, also averaged per cell, helps differentiate various terrain types (e.g., trees, vegetation, barren land) based on surface characteristics. 

We also work with Digital Elevation Maps(DEMs) for areas where LiDAR data is not available. The DEMs are transformed into a 2.5D point cloud, which is based on the resolution of the scan. After this conversion, we process the data similarly to how we handle LiDAR data, to create the necessary inputs for Trailblazer.

\subsubsection{SegFormer}
SegFormer \cite{xie2021segformer}, a lightweight transformer-based architecture for semantic segmentation, is utilized to generate a 2D pixel-wise mask of the overhead image, classifying each pixel into distinct semantic classes. Segformer features a hierarchically structured transformer encoder coupled with a lightweight MLP decoder that effectively integrates both local and global attentions.

To train the SegFormer model, we use the FLAIR-one dataset \cite{ign2022flair1}, which provides semantic annotations for aerial imagery along with topographic and land cover information. This dataset consists of 77,412 high-resolution patches, each measuring 512×512 pixels with a spatial resolution of 0.2 meters, spanning 19 distinct semantic classes.
\begin{table}[ht]
\small
\centering
\renewcommand{\arraystretch}{1.5}
\resizebox{\textwidth}{!}{%
\begin{tabular}{|c|p{8cm}|c|}
\hline
\textbf{Category}       & \textbf{FLAIR Labels}                                    & \textbf{Class IoU (\%)} \\ \hline
\textbf{Traversable}    & Herbaceous region, Agricultural, Plowed land, Bare soil & 71.88                 \\ \hline
\textbf{Non-Traversable} & Pervious surface, Impervious surface, Ligneous, Mixed    & 52.96                 \\ \hline
\textbf{Vegetation}     & Coniferous, Deciduous, Brushwood, Vineyard, Clear cut     & 73.25                 \\ \hline
\textbf{Obstacles}      & Building, Greenhouse, Other                               & 77.88                 \\ \hline
\textbf{Water}          & Water, Swimming pool                                     & 82.50                 \\ \hline
\multicolumn{2}{|c|}{\textbf{Mean IoU (mIoU)}}                                      & \textbf{71.69}        \\ \hline
\end{tabular}%
}
\caption{FLAIR Label Categories with Class IoU Metrics}
\label{tab:flair_labels}

\end{table}

We preprocess the FLAIR-one annotations by consolidating the 19 semantic classes into 5 superclasses, tailored to off-road terrains where fewer distinguishing features are typically required. This grouping is based on the semantic relevance of the classes within the environmental context. The resulting superclasses are outlined in Table \ref{tab:flair_labels}.

\subsection{Trailblazer Architecture}
The Trailblazer architecture comprises two primary components: an encoder-decoder framework and a differential A* planner. The encoder-decoder architecture utilizes fully convolutional layers in conjunction with attention mechanisms to effectively extract features from the input data.

\subsubsection{Encoder-Decoder:}
\begin{figure}[ht]
    \centering
    \includegraphics[width=\textwidth]{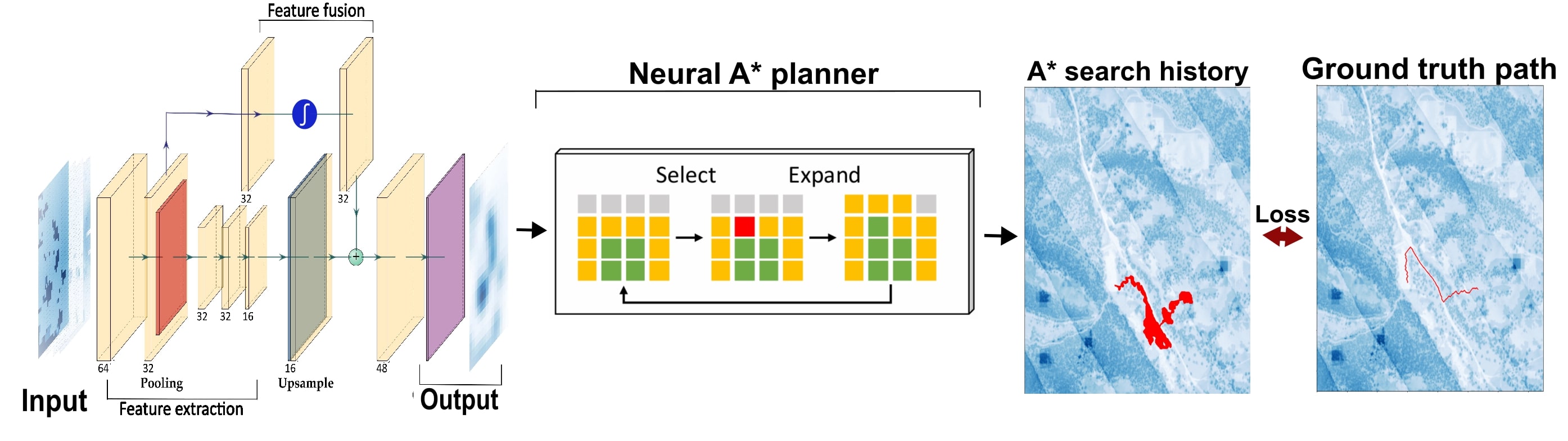}
    \caption{Trailblazer framework architecture. The backbone consists of feature fusion and feature extraction branch generating the costmap. The Neural A* planner generates the path from the costmap compared against ground truth to compute loss.}
    \label{fig:Trailblazer_arch}
    \vspace{-0.5 cm}
\end{figure}

Inspired by multiscale convolutional networks \cite{doi:10.1177/0278364917722396}, Trailblazer integrates feature extraction and fusion modules. The feature extraction module comprises four convolutional layers and a pooling layer (Figure \ref{fig:Trailblazer_arch}), enabling effective low-level feature sharing. To retain both translationally variant and invariant features, a spatial attention module \cite{10.1007/978-3-030-01234-2_1} is incorporated at the skip connection. The module's output is upsampled and concatenated with the skip connection, allowing independent treatment of feature channels. A final convolution on the concatenated output produces the cost map.

\subsubsection{Neural Astar Planner:}
We integrate the Neural A* algorithm \cite{pmlr-v139-yonetani21a} into our neural network architecture to compute paths between start and goal points on a cost map. Neural A* is a data-driven search method that redefines the traditional A* algorithm to be differentiable, enabling seamless integration with a convolutional encoder to form an end-to-end trainable neural network planner. The differentiable A* module employs techniques like discretized activation inspired by Hubara et al. \cite{NIPS2016_d8330f85}, allowing it to execute A* searches during the forward pass and back-propagate losses through all search steps to other trainable backbone modules. The loss function used is the mean L1 loss between the search history and the ground-truth path. This integration enhances the network's ability to learn from expert demonstrations, producing paths that closely align with ground-truth trajectories in terms of accuracy and efficiency.

\section{Experiments}
\subsection{Semantic Segmentation}
\begin{figure}[ht]
    \centering
    \begin{minipage}[t]{0.48\textwidth}
        \centering
        \includegraphics[width=\textwidth, height = 4.5 cm]{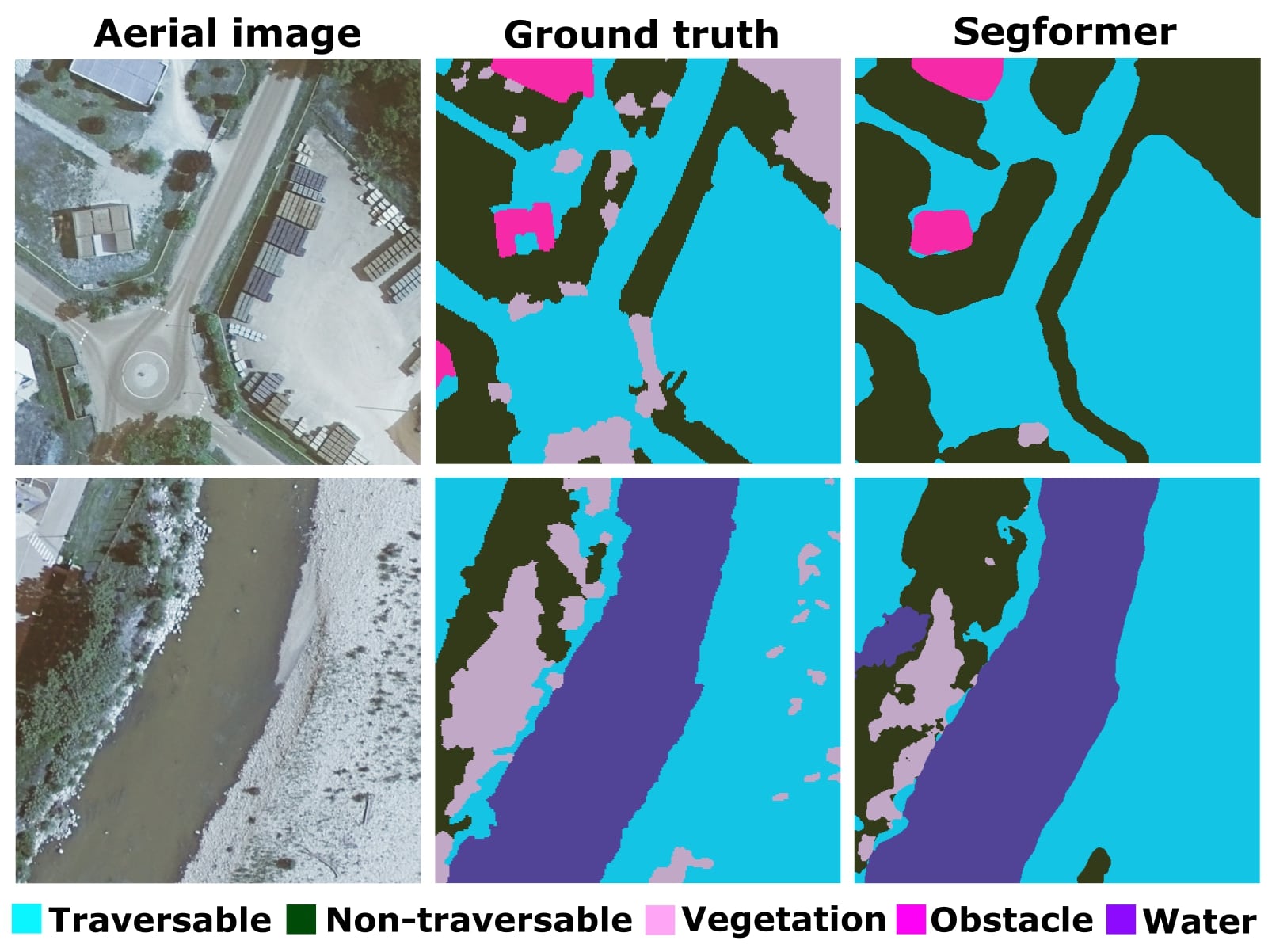}
        \caption{Semantic ground truths and predictions from Segformer.}
        \label{fig:segformer}
    \end{minipage}%
    \hfill
    \begin{minipage}[t]{0.48\textwidth}
        \centering
        \includegraphics[width=\textwidth, height =  4.5 cm]{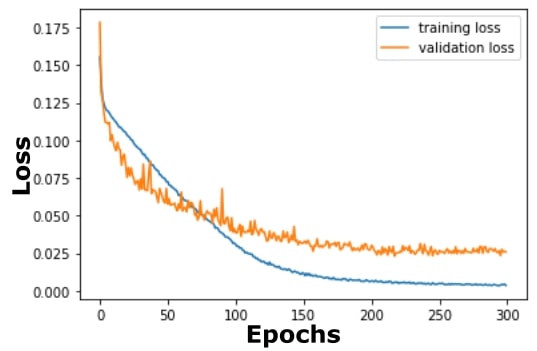}
        \caption{Training and validation loss curves of Trailblazer.}
        \label{fig:training_curve}
    \end{minipage}
\end{figure}

We trained the Segformer model on the FLAIR-One dataset using a modified superclass annotation strategy to improve segmentation performance. By grouping classes into superclasses, we effectively reduced the complexity of the confusion matrix, leading to improved accuracy. The dataset was divided into 61,710 images for training and 15,700 images for validation. The model was trained for 400 epochs using a categorical cross-entropy loss function. This training process resulted in the model achieving a mIoU of 87.5\% and an accuracy of 71.69\% on the validation set. Examples of predictions on the validation set are illustrated in Figure \ref{fig:segformer}. 
\subsection{Trailblazer}
The Trailblazer model was initially trained using simulation data and later evaluated on real-world trajectories. For data collection, we employed the MAVS off-road simulator \cite{7988748} to navigate diverse scenes and generate trajectory data. A total of 4,000 data instances were collected, comprising input maps and their corresponding trajectories. This dataset was split into training and validation subsets to train the Trailblazer model effectively. The training and validation loss curves, shown in Figure \ref{fig:training_curve}, demonstrate that the model achieved convergence, with a minimum validation loss of 0.0256. When tested on a separate dataset, the model achieved a loss of 0.0213, indicating its effectiveness in trajectory prediction.

\begin{figure}[!ht]
    \centering
    \includegraphics[width=\textwidth]{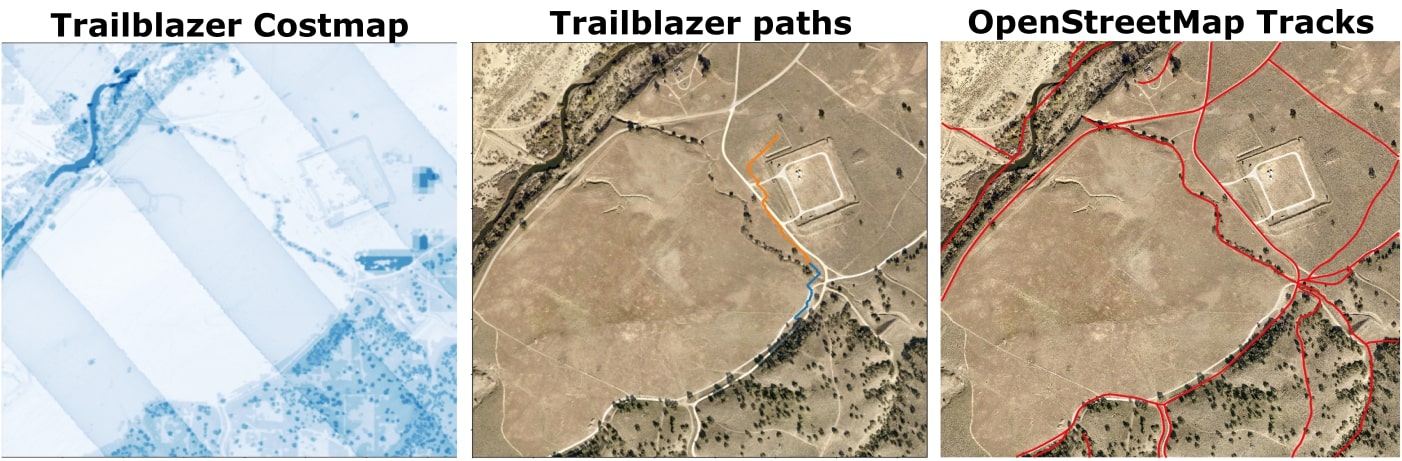}
    \caption{Costmap, paths extracted from costmap and OpenStreetmaps tracks from a test site.}
    \label{fig:trails}
\end{figure}

To enable sim-to-real transfer, we retrained the simulation-trained Trailblazer model using real-world trajectory data. OpenStreetMap (OSM) \cite{OpenStreetMap} provides GPS trajectories of vehicles driven across various locations, which we leverage to further train Trailblazer for enhanced generalization across diverse geographic regions. These trajectories, exemplified in Figure~\ref{fig:trails}, serve as valuable data to improve the model's adaptability.The model was evaluated on various real-world overhead datasets obtained from the USGS, with an example illustrated in Figure~\ref{fig:trails}.

\subsubsection{Real-world Experiments:}
Real-time navigation experiments were conducted at sites in North Carolina in association with DARPA RACER program, where Trailblazer functions as a global planner in conjunction with a local planner.
\vspace{-0.5 cm}
\begin{figure}[ht]
    \centering
    \includegraphics[width=\textwidth]{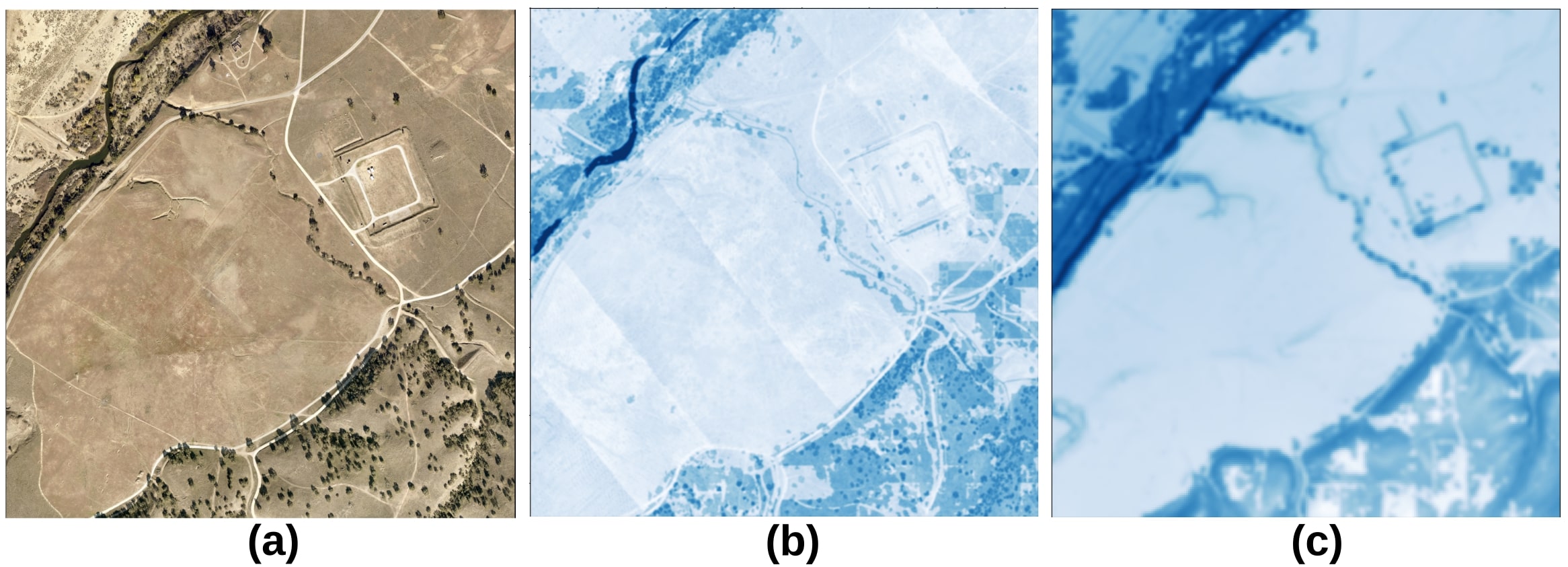}
    \caption{LiDAR vs DEM: (a) Satellite image from a test site (b) Costmap generated from LiDAR data with 4 $points/m^2$ resolution., (c) Costmap generated from DEM with a $1/3 arc-sec$ resolution. }
    \label{fig:lidvdem}
\end{figure}
\section{Experimental Insights}

Traditional path planning relied on manual data collection via UAVs, limiting use in controlled/ restricted air spaces. Our work shows open-source data (USGS, ESRI, OpenStreetMap) enables robust urban/off-road navigation without costly manual efforts. Trailblazer achieves cross-terrain adaptability using these datasets, proving open-source data is sufficient for scalable global planning.

Trailblazer also includes capbility to generate costmaps from Digital Elevation Maps(DEMs) replacing 3D LiDAR data. While DEMs produced costmaps of comparable quality to 3D LiDAR, their effectiveness depended heavily on resolution. DEMs with resolutions below 1/9 arc-second resulted in subpar costmaps and slope distortions. Figure \ref{fig:lidvdem} compares the quality of costmaps generated from LiDAR and DEMs. Since LiDAR data isn't globally available, we aimed to utilize widely mapped DEMs. 

While handling data from different sources, one must be careful when transforming coordinates between various datums. An incorrect transformation can lead to misalignment. This misalignment can result in inaccurate projections and costmaps that do not line up properly.

Trailblazer started as a local planning algorithm\cite{vishwanath2022camellearningcostmapseasy} designed for navigating off-road terrains. Over time, it has evolved to handle not just local planning, but also global planning. This advancement is made possible by the algorithm's ability to dynamically update the global costmap with real-time data collected onboard. For instance, consider a situation where overhead data may indicate the presence of a dry riverbed as a viable path. However, a local inspection might show that this specific area is, in fact, not traversable. In such cases, the global costmap can adjust itself based on the latest information, which enables Trailblazer to determine the best trajectory possible. This feature fosters strong and effective planning in environments that are constantly changing.

\section{Conclusion}
In this paper, we presented Trailblazer, a novel imitation learning framework designed for long-range planning in off-road terrains. By leveraging multi-modal sensor data, including satellite imagery and LiDAR, Trailblazer generates costmaps for efficient path planning without the need for manual tuning. Extensive real-world testing validates Trailblazer’s robust performance, demonstrating its potential for scalable and efficient autonomous navigation. The framework also showcases the utility of open-source data from sources like USGS, ESRI, and OpenStreetMap, proving that such data is sufficient for robust urban and off-road navigation.

Trailblazer's ability to generate costmaps from Digital Elevation Maps (DEMs) offers a viable alternative when LiDAR data is unavailable, although the effectiveness depends on the resolution of the DEMs. Looking ahead, we plan to adjust our framework so that it can utilize high-resolution satellite images alongside low-resolution Digital Elevation Maps. Our goal is to create costmaps that are similar in quality to those made with LiDAR data. This approach will allow us to produce costmaps using data that is more widely accessible all over the world.

\section{Acknowledgments}
Research was sponsored by DEVCOM Army Research Laboratory (ARL) and was accomplished under Cooperative
Agreement Number W911NF-21-2-0064. The views and conclusions contained in this document are those of the
authors and should not be interpreted as representing the official policies, either expressed or implied, of the ARL or
the U.S. Government. The U.S. Government is authorized to reproduce and distribute reprints for Government
purposes notwithstanding any copyright notation herein.
\bibliographystyle{splncs03_unsrt}
\bibliography{bibliography}

\begin{thebibliography}{10}
\providecommand{\url}[1]{\texttt{#1}}
\providecommand{\urlprefix}{URL }

\bibitem{Vandapel-2003-8764}
Vandapel, N., Donamukkala, R.R., Hebert, M.: Quality assessment of traversability maps from aerial lidar data for an unmanned ground vehicle. In: Proceedings of (IROS) IEEE/RSJ International Conference on Intelligent Robots and Systems. vol.~1, pp. 305 -- 310 (October 2003)

\bibitem{machines11080807}
Sánchez, M., Morales, J., Martínez, J.L.: Waypoint generation in satellite images based on a cnn for outdoor ugv navigation. Machines  11(8) (2023)

\bibitem{6284849}
High Performance Outdoor Navigation from Overhead Data using Imitation Learning, pp. 262--269 (2009)

\bibitem{hdif2023}
Guaman~Castro, M., Triest, S., Wang, W., Gregory, J.M., Sanchez, F., Rogers~III, J.G., Scherer, S.: How does it feel? self-supervised costmap learning for off-road vehicle traversability (2023)

\bibitem{9812238}
Weerakoon, K., Sathyamoorthy, A.J., Patel, U., Manocha, D.: Terp: Reliable planning in uneven outdoor environments using deep reinforcement learning. In: 2022 International Conference on Robotics and Automation (ICRA). pp. 9447--9453

\bibitem{huang23vlmaps}
Huang, C., Mees, O., Zeng, A., Burgard, W.: Visual language maps for robot navigation. In: Proceedings of the IEEE International Conference on Robotics and Automation (ICRA). London, UK (2023)

\bibitem{USGSLidarExplorer}
{U.S. Geological Survey}: {USGS Lidar Explorer} (2025), \url{https://apps.nationalmap.gov/lidar-explorer/}

\bibitem{arcgis}
Esri: Arcgis geographic information system (2025), version 11.3. Available at: \url{https://www.esri.com/arcgis}

\bibitem{xie2021segformer}
Xie, E., Wang, W., Yu, Z., Anandkumar, A., Alvarez, J.M., Luo, P.: Segformer: Simple and efficient design for semantic segmentation with transformers. In: Neural Information Processing Systems (NeurIPS) (2021)

\bibitem{ign2022flair1}
Garioud, A., Peillet, S., Bookjans, E., Giordano, S., Wattrelos, B.: Flair 1: semantic segmentation and domain adaptation dataset  (2022), \url{https://arxiv.org/pdf/2211.12979.pdf}

\bibitem{doi:10.1177/0278364917722396}
Wulfmeier, M., Rao, D., Wang, D.Z., Ondruska, P., Posner, I.: Large-scale cost function learning for path planning using deep inverse reinforcement learning. The International Journal of Robotics Research  36(10),  1073--1087 (2017)

\bibitem{10.1007/978-3-030-01234-2_1}
Woo, S., Park, J., Lee, J.Y., Kweon, I.S.: Cbam: Convolutional block attention module. In: Ferrari, V., Hebert, M., Sminchisescu, C., Weiss, Y. (eds.) Computer Vision -- ECCV 2018. pp. 3--19. Springer International Publishing, Cham (2018)

\bibitem{pmlr-v139-yonetani21a}
Yonetani, R., Taniai, T., Barekatain, M., Nishimura, M., Kanezaki, A.: Path planning using neural a* search. In: Meila, M., Zhang, T. (eds.) Proceedings of the 38th International Conference on Machine Learning. Proceedings of Machine Learning Research, vol. 139, pp. 12029--12039. PMLR (18--24 Jul 2021)

\bibitem{NIPS2016_d8330f85}
Hubara, I., Courbariaux, M., Soudry, D., El-Yaniv, R., Bengio, Y.: Binarized neural networks. In: Lee, D., Sugiyama, M., Luxburg, U., Guyon, I., Garnett, R. (eds.) Advances in Neural Information Processing Systems. vol.~29 (2016)

\bibitem{7988748}
Hudson, C., Goodin, C., Miller, Z., Wheeler, W., Carruth, D.: Mississippi state university autonomous vehicle simulation library. In: Proceedings of the Ground Vehicle Systems Engineering and Technology Symposium. pp. 11--13 (2020)

\bibitem{OpenStreetMap}
{OpenStreetMap contributors}: {Planet dump retrieved from https://planet.osm.org }. \url{ https://www.openstreetmap.org } (2017)

\bibitem{vishwanath2022camellearningcostmapseasy}
Vishwanath, K., Sujit, P.B., Saripalli, S.: Camel: Learning cost-maps made easy for off-road driving (2022), \url{https://arxiv.org/abs/2209.12413}

\end{thebibliography}

\end{document}